\documentclass[a4paper,12pt]{article}
\usepackage{color,soul}
\usepackage[english]{babel}
\usepackage[utf8]{inputenc}
\usepackage{rotating}
\usepackage{tikz}
\usepackage{amsmath}
\usepackage{graphicx}
\usepackage[colorinlistoftodos]{todonotes}
\usepackage[margin=1in]{geometry}
\usepackage{hyperref}
\usepackage{amssymb}
\usepackage{stix}
\usepackage{gensymb}
\usepackage{adjustbox}      
\usepackage{graphicx} 
\usepackage[bf]{caption}
\usepackage[singlelinecheck=false, aboveskip=-3pt]{subcaption}

\usetikzlibrary {positioning}

\title{ Combining Discrete Choice Models and Neural Networks through Embeddings: Formulation, Interpretability and Performance }

\bigskip

\usepackage{float}
 \author{\textit{Ioanna Arkoudi$^{1*}$, Carlos Lima Azevedo$^1$, Francisco C. Pereira$^1$}
 \\
\textit{$^1$DTU Technical University of Denmark}
\\
\small{
\textit{$^*$ corresponding author}, \href{ioaar@dtu.dk}{ioaar@dtu.dk}
}}

\begin{document}
\maketitle
\thispagestyle{empty}

\begin{abstract}
This study proposes a novel approach that combines theory and data-driven choice models using  Artificial Neural Networks (ANNs). In particular, we use continuous vector representations, called embeddings, for encoding categorical or discrete explanatory variables with a special focus on interpretability and model transparency. Although embedding representations within the logit framework have been conceptualized by Pereira in \cite{camara},   their dimensions do not have an absolute definitive meaning, hence offering limited behavioral insights in this earlier work. The novelty of our work lies in enforcing interpretability to the embedding vectors by formally associating each of their dimensions to a choice alternative.
Thus, our approach brings benefits much beyond a simple parsimonious representation improvement over dummy encoding, as it provides behaviorally meaningful outputs that can be used in travel demand analysis and policy decisions. Additionally, in contrast to previously suggested ANN-based Discrete Choice Models (DCMs) that either sacrifice interpretability for performance or are only partially interpretable, our models preserve interpretability of the utility coefficients  for all the input variables despite being based on ANN principles. The proposed models were tested on two real world datasets and evaluated against benchmark and baseline models that use dummy-encoding. The results of the experiments indicate that our models deliver state-of-the-art predictive performance, outperforming existing ANN-based models while drastically reducing the number of required network parameters. 
\end{abstract}
\newpage
\tableofcontents
\newpage
\section{Introduction}

Discrete choice models (DCMs) based on random utility maximisation (RUM) constitute the typical theoretical framework for modeling traveling behavior. The logit model family is the most commonly used for choice modeling, where models are generally formulated with parametric utility functions that describe individual preferences and are a known source of valuable information for demand modeling and forecasting \cite{cirillo}. One of the central tasks in choice modeling is to find a model specification, particularly regarding variable selection and representation (e.g. discretization, transformation) that best describe the decision process of an individual in a given context. Accomplishing this task poses implications regarding efficient survey design and sampling, since the data requirements for producing reliable and robust parameter estimates, is a complex and multifaceted issue. Many interacting factors such as sample size, number of considered alternatives, number and levels/range of attributes come into play that can greatly affect the quality of the model's results \cite{rose}.  

Given that travel data collection is considered the most expensive and time consuming part of the transportation model development process \cite{kulpa}, it is not surprising that small sample size has been recognised as one of the major challenges for model estimation \cite{challenges}. At the same time, including richer data has been identified as a crucial task for promoting future transport research \cite{challenges}, which however increases further the data size requirements and model complexity. One way to address such challenges is to find new variable representation paradigms that more compactly and sufficiently capture population heterogeneity, while being more robust to the relatively limited amount of data that is usually available in travel surveys.

Since their early days, variable representation in logit models has followed generally quite clear principles. Numeric quantities like distance, travel time, cost or income may be directly used in a model - or transformed depending on observed non-linear effects (e.g. using logarithmic transformations). Variables that are not numerical, i.e. ordinal and nominal variables, should be encoded into numerical values using encoding techniques, since the models support only numerical attributes as input. The default encoding method for non-numeric variables used in logit models is dummy encoding, also known as "one-hot-encoding" in machine learning literature. Typical cases of variables that require such treatment in transport choice modeling are numeric variables, such as age, income or even geographic coordinates which tend to be discretized. Similarly, categorical variables such as education level or trip purpose are already discrete, and thus are also usually "dummyfied".
 
 One-hot encoding, although simple and commonly used, has several inherent disadvantages. As it adds a variable for each unique variable category, the  model's dimensionality increases proportionally to the cardinality of the categorical variables considered. In cases where the cardinality of the categorical variables is large, we end up with a high-dimensional model representation which may be subject to what is infamously known as "the curse of dimensionality” introduced in \cite{bellman}, that poses severe challenges in statistical modeling as an increase in the the number of dimensions requires an exponential increase in the sample size for the model to be reliable \cite{giron,monte_carlo}.

In relatively small data sets,
it may lead to overfitting, that is a common issue for models with high complexity \cite{hughes}. Overfitting describes the situation when the model is sensitive to random errors and noise in the given data instead of learning the underlying relationship between the dependent and independent variables. Such models typically tend to demonstrate high goodness-of-fit in-sample but are unstable and yield poor performance in out-of-sample predictions, which renders them ill-suited for projecting accurate future demand and quite questionable on whether they accurately explain the behavior they are set to represent. In contrast to machine learning, where focus is placed on model’s generalization performance on out-of-sample data, in classical econometric methods, model selection typically relies on comparing the in-sample goodness of fit across specifications \cite{mullain}. Consequently, potential overfitting remains undetected, and the interpretation of the model’s coefficients and computed elasticities used for planning and policy-making may be subject to errors.

The "curse of dimensionality" also manifests itself in datasets where behavior heterogeneity is high, and the number of observations is significantly smaller than population size. In such cases, increased dimensionality is very risky because the amount of data that contributes to estimating each coefficient becomes insufficient. A simple intuition here is by considering that, for a dummy variable that is only "1" for a few observations in the dataset, its coefficient will only be "activated" that small number of times. If there is a lot of variance in the associated behavior, the variance of the coefficient will also be large, and the coefficient will be considered statistically insignificant.


In addition to the above-mentioned practical problems related to data and estimation requirements, another issue with one-hot encoding is that the variable categories are assumed to be mutually exclusive and unrelated, creating representations that are forced to be orthogonal and equidistant. Such representations fail to capture the similarity between categories that would be intuitive for a human, as they ignore their associative or semantic relatedness. For example, when looking at a variable that indicates a person's occupation containing the following categories: Pension, Self-Employed, Employee, Unemployed, Student, HighSchool, PrimSchool, a human observer could infer, that the last 3 categories are more similar/related to each other as they all refer to a person being under education compared to Employee or Self-Employed i.e. being a worker. However a model that uses one-hot encoding to represent a categorical feature does not capture such contextual information, and thus cannot make such distinctions. Operationalizing the data in a way that more closely resembles human perception and intuitions is a crucial component for creating realistic behavioral models that actually reflect the decision making processes underlying the observed (or stated) behavior of an individual.

The key idea in this paper is to augment traditional DCMs by incorporating a data-driven method of encoding categorical variables using continuous vector representations, called \emph{embeddings}. This method aims to mitigate the aforementioned problems of one-hot encoding and is tailored to the problem at hand as it follows a supervised approach for learning the representations. It is strongly inspired by natural language processing (NLP) techniques, namely the word2vec algorithm \cite{WORD2VEC} that is based on artificial neural networks (ANN). Thus, the current work contributes to the body of research that aims to combine theory and data-driven choice models. Although our models, referred to as the Embeddings Multinomial Logit (E-MNL) and Embeddings Learning Multinomial Logit (EL-MNL) models respectively, are primarily developed to be used within the context of travel choice behaviour, they provide a general flexible framework that can find applications outside the Transport field.We also note that the source code is openly available \footnote{\url{https://github.com/ioaar/Interpretable-Embeddings-MNL}} together with a detailed Jupiter (iPython) notebook tutorial that provides further explanation regarding the proposed models and example code to reproduce in other applications.

In contrast to previously suggested ANN-based DCMs, that either compromise interpretability for performance or are only partially interpretable, our models combine transparency and straightforward interpretability with predictive power. They avoid overfitting in in-sample data and show higher out-of-sample predictive performance on 2 real-world datasets when compared to benchmark models, allowing for more reliable model interpretation and more accurate/realistic demand forecasting. Their particular architecture further allows us to generate continuous vector representations of the input categorical variables that combine many desirable properties, namely compact dimensionality, incorporating  their semantic-relatedness, and being tailored-made for the specific behavior we aim to model. Their estimation comes with a relatively low computational cost as they, similarly to word2vec, follow a shallow neural architecture, which limits the training time demands without compromising the quality of the resulting representations. Due to their compact size, embeddings encoding allows us to handle explanatory variables of notably high cardinality and sparsity, such as  POIs, departure times, and origin and destination variables that are commonly ignored in choice modelling despite their potential availability. What is more, the embedding representations presented in this study possess certain useful intrinsic properties such as interpretability, and resusability. The former enables us to acquire meaningful  representations of the categorical variables in a continuous space, which can be used for data visualization and analysis, amongst other purposes. The latter allows for reusing the categorical  embeddings learned from one dataset for other ones, as long as the study behavior keeps consistent between the datasets. 
 
\section{Related Work}

\subsection{Combining ANNs and DCMs}

Over the recent years, a growing number of research studies is devoted to exploring deep learning approaches and ANNs for enchancing DCMs, aiming to blend theory-based and data-driven approaches. Despite their high prediction performance, ANN-based models have been often criticized to be a "black box": their inner workings are difficult to understand, and their parameters do not provide any straightforward insights about how the input variables affect the predicted output. Recent studies in the transport field have paid attention to this issue, aiming to create ANN-based DCMs that not only perform better than traditional DCMs but also provide interpretable and behaviorally meaningful outputs that can be used in travel demand analysis and policy decisions.

Several of these studies \cite{wang_a, han,sin} focus on discovering the optimal utility function with the use of ANNs that allow them to augment standard DCMs with a flexible non-linear component able to capture systematic heterogeneity without posing assumptions on its origin. In these studies, the proposed models increased the overall predictive performance compared to standard DCMs, showing promising results, while keeping the models' parameters, at least partially, interpretable. 

Of this series of studies, the work more relevant to ours is that of Sifringer et al. \cite{sin}. Based on previous works \cite{bentz, hruschka_1,hruschka_2}
that investigated the potential of ANN-based choice models, they developed two hybrid models, referred to as Learning Multinomial Logit (L-MNL) and Learning Multinomial
Nested Logit (L-NL) respectively. They consist of a linear layer that represents the systematic part of the utility, and a non-linear densely connected layer aiming to learn a representation term from a set of extra input explanatory variables for which no a priori relationship is assumed. The proposed models outperformed the standard DCMs and existing hybrid models on real and synthetic datasets, while they were designed to keep the linear layer parameters interpretable, for use in choice analysis.


Lastly, it is worth-mentioning the work of Wong \& Farooq \cite{resnet} that used the concept of residual learning, originally introduced by He et al. \cite{he}, to develop  a residual network (ResNet) that integrates a Deep Neural Network (DNN) architecture into an MNL. The proposed model, referred to as ResLogit model, showed promising performance when compared to an MNL and a Multilayer Perceptron on a large dataset, and produced lower standard errors for the estimated utility parameters. Additionally, ResLogit's architecture allows for increased interpretability as the matrix of residual network's parameters can be used for extracting meaningful econometric indicators.

Our work differs from the aforementioned studies in that we propose a novel ANN-based DCM with fully interpretable parameters. This is achieved by introducing an embedding layer as part of the model's architecture that can effectively encode discrete input variables with high cardinality into alternative-specific continuous values. The main model, i.e. E-MNL, presented in this study is fully transparent and behaviorally interpretable while at the same time it aims to increase the overall predictive performance using a minimal number of network parameters. Similarly to Sifringer et al. \cite{sin} we provide a model framework that estimates jointly the linear (utility) parameters and the non-linear ones, however it is able to provide utility coefficients and post-estimation statistics for all the input variables, not only a subset of them. Additionally the generated embedding representations are able to reveal data-driven insights regarding latent information and patterns residing in the data that can be used for diagnostic and analysis purposes. Lastly, in an effort to further increase predictive performance while preserving interpretability, we develop EL-MNL by extending the architecture of E-MNL to incorporate the one suggested in Sifringer et al. \cite{sin} while maintaining meaningful utility parameters for all the input variables.

\subsection{Embedding representations}

The popularization of embedding representations can be attributed to the creation of word2vec \cite{WORD2VEC}, a deep learning-based method for generating dense vectors of words, known as word embeddings. Word embeddings alleviate the problems related to high-dimensionality and sparsity of language data and offer compact and intuitive representations that manage to capture the semantic relatedness of the encoded units based on their distributional properties \cite{ngrams}. They have proven to be highly effective at a wide variety of downstream NLP tasks ranging from text classification and question answering to automatic machine translation \cite{downstream1, downstream2}. 


Over the past years, numerous studies have demonstrated that ANN-embedding methods of representing discrete informational units can also find applications outside the field of NLP. Indicatively we mention Node2Vec \cite{node2vec} and DeepWalk \cite{deepwalk} for representing graphs, TransE \cite{transe} and RESCAL \cite{rescal} for representing knowledge graphs, as well as DeViSE \cite{devise} that maps images into a rich semantic embedding space. Several studies \cite{place2vec,urban2vec,poi2vec,geoteaser} focused on generating place/geospatial embedding representations in a similar manner to word2vec taking into account mobility patterns and spatial context information. The resulting representations offered new insights into the understanding of space semantics and place functionalities \cite{place2vec, urban2vec} and found applications in various tasks related to urban planning and policy design such as point-of-interest (POI) recommendation \cite{geoteaser}, and predicting users who will visit a given POI in a given future period \cite{poi2vec}.

In studies more relevant to transport and demand forecasting applications, de Brébisson et al. trained multiple ANN models of varying architectures to predict the destination of a taxi \cite{taxi} based on the beginning of its trajectory and associated metadata. The results showed that learning embeddings jointly with the models to encode the discrete meta-data (client ID, taxi ID, date
and time information) significantly improved
the predictions of taxi destinations. Guo \& Berkhahn \cite{guo} also generated ANN-based categorical embeddings and succesfully applied this encoding method for forecasting purposes. They demonstrated that using the embeddings as input features in different ML algorithms considerably improved the daily sales forecasting performance, and showed that embeddings encoding helps the NN to generalize better when the data is sparse and statistics unknown, where other methods tend to overfit. More recently, Wang et al. \cite{wang_emb} investigated the application of discrete variables in traffic prediction models based on NNs, considering a large amount of categorical data, such as time, site ID and weather. They compared the embedding representations of such variables to one-hot encoding and showed that the embedding vectors are better able to represent the internal variable relationships, and thus are more efficient in predicting traffic flow. Additionally, they showed that the intrinsic properties of the categorical variables and the relationship between them can be revealed through visual analysis of the trained embedding vectors, which can further be used as a potential supervised clustering method. In a similar study, dedicated to bike-sharing demand forecasting, Ahn et al. \cite{ahn} proposed a categorical embedding based ANN-model, using techniques such as Batch Normalization, Dropout, and Cyclical Learning to avoid overfitting. Their model outperformed the traditional one-hot encoding based ones, while performing dimensionality reduction and clustering on the trained embedding vectors proved to be useful for effectively analysing the demand patterns.

In a work more relevant to ours, Pereira \cite{camara} presented a method of mapping discrete variables, that are typically used in travel demand modeling, into a latent embedding space. He further demonstrated that using the embedding encoding for categorical variables in a choice model outperforms traditional methods of data representations, such as dummy variables or PCA encoding in out-of-sample evaluation.


In all the aforementioned studies, any attempt to interpret the embedding space of the categorical data is limited to comparing the relative distances between the embedding vectors, i.e. categories that have a similar effect on the target variable will tend to be close to each other in the embedding space. Despite the insights uncovered from such embedding representations and their potential usefulness for visualization and clustering purposes, neither the embedding vectors nor their dimensions have an absolute, definitive meaning. Additionally, the optimal number of embeddings dimensions used for encoding categorical variables is usually not known beforehand and is subject to tuning, typically spanning to few tens of dimensions.

Our work is novel in this respect as we enforce interpretability to the embedding vectors by formally assigning meaning to their dimensions. More specifically, we define the learning algorithm such that each embedding dimension corresponds to a choice alternative, and thus, during training, the embedding vectors of the categories that are more relevant for choosing a specific alternative receive larger values along its corresponding dimension. Such an approach allows us not only to create an ANN-embedding model with fully interpretable parameters and embedding representations with meaningful dimensions, but also to minimize the dimensionality of the embedding space by restricting it to the number of choice alternatives without sacrificing predictive performance.

\section{Model Architecture}

\subsection{MNL as an ANN} \label{MNL_as_ANN}
Given a choice set $C$ with $J$ number of alternatives, we consider a multinomial choice model, where  and $X=\{ x_{1}, x_{2}, ..., x_{K}\}$ are the explanatory variables representing the observed attributes of the choice alternatives and the individual’s socio-demographic characteristics.
The utility that individual $n$ associates with alternative $i=1..J$ is formally given by:

\begin{equation} \label {eq1}U_{i,n}= V_{i,n} + \epsilon \end{equation} where $\epsilon$ is independently and identically distributed Type I Extreme Value.
\newline Assuming that the systematic part of the utility is linear-in-parameter and conveniently considering a single vector of coefficients that applies to all the utility functions, $V_{i,n}$ can be described by the following equation:

\begin{equation} \label {eq2}V_{i,n}=BX_{i,n}\end{equation} where $B=\{ \beta_{1}, \beta_{2}, ..., \beta_{K}\}$  are the  preference parameters associated with the $K$ explanatory variables in the vector $X_{i, n}$ corresponding to explanatory variables for individual $n$.

The current study adopts the implementation of an MNL as an ANN using a simple 2D-Convolutional Neural Network (CNN) architecture as in \cite{sin}, which will be briefly explained in this section. Although CNNs are traditionally used to analyse image and signal data using complex architectures that typically include non-linear activation functions and multiple channels and convolution layers, their weight-sharing architecture conveniently allow us to use them in a more simplified form to retrieve the MNL formulation as defined in (\ref {eq2}).  
\newline In order to explain the formulation according to \cite{sin}, we need to re-arrange the input space $X$ as a 2-dimensional matrix of shape $J \times K $, and define a convolution filter consisting of an array of trainable weights $B$ of shape $1 \times K$. We then consider a CNN, with a single convolution layer. The model maps $X$ to an output space $V$ by sliding across the $k=1..K$ variables of the input $X$, one row (i.e. alternative $i$) at a time, and applying the dot product between $X$ and $B$, yielding each time a single scalar value and resulting in the output space $V$ of shape $(J, 1)$, such that:
\begin{equation} \label {eq3}V_{i}= \sum_{k=1}^{K} B_{k}(X_{k})_{i}\end{equation}

We can thus observe that this convolution process is equivalent to the utility functions $V_n = \{V_{1,n},..., V_{J,n}\}$ as defined in equation \ref{eq2}. After the convolution takes place, the output $V$ representing the utilities is propagated to the final activation layer, which generates the probability distribution over $J$ different choice alternatives using the softmax activation function $\sigma$, such that:

\begin{equation}\label {eq4}
\left(P_n\right)_{i}= \left(\boldsymbol{\sigma}\left(\mathbf{v}_{n}\right)\right)_{i}=\dfrac{e^{v_{i,n}}}{\sum_{j=1}^{J} e^{v_{j,n}}}\end{equation}
which is equivalent to the probability for individual $n$ to select choice alternative $i$ within the MNL framework under the standard assumptions. 

\begin{figure}[H]
\includegraphics{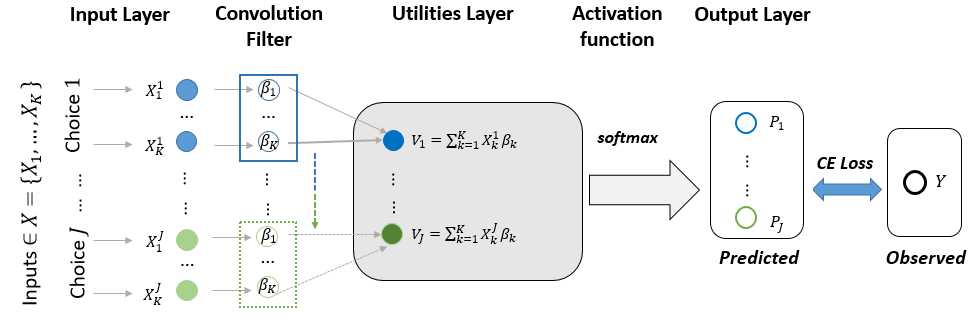}
\centering
\caption{\textit{Network architecture implementing an MNL as a 2D-CNN.}}
\label{fig:mnl_simple}
\end{figure}

As is usually the case when the output layer activation function of an ANN is softmax, cross entropy is used as a loss function to optimize the model's parameters, i.e. the weights of $B$, during training through backpropagation. As noted in \cite{sin}, minimizing cross entropy loss is equivalent to maximizing the log-likelihood function, and thus allows us to derive the parameters’ Hessian  matrix and compute useful post-estimation indicators such as their standard errors and confidence intervals of the model. The architecture is shown in Figure \ref{fig:mnl_simple}.
\subsection{Interpretable Embeddings MNL (E-MNL)}

\subsubsection {General formulation and model architecture}
\begin{sidewaysfigure}
\vspace*{5in}
    \begin{tikzpicture}\includegraphics[width=\textwidth,height=\textheight,keepaspectratio]{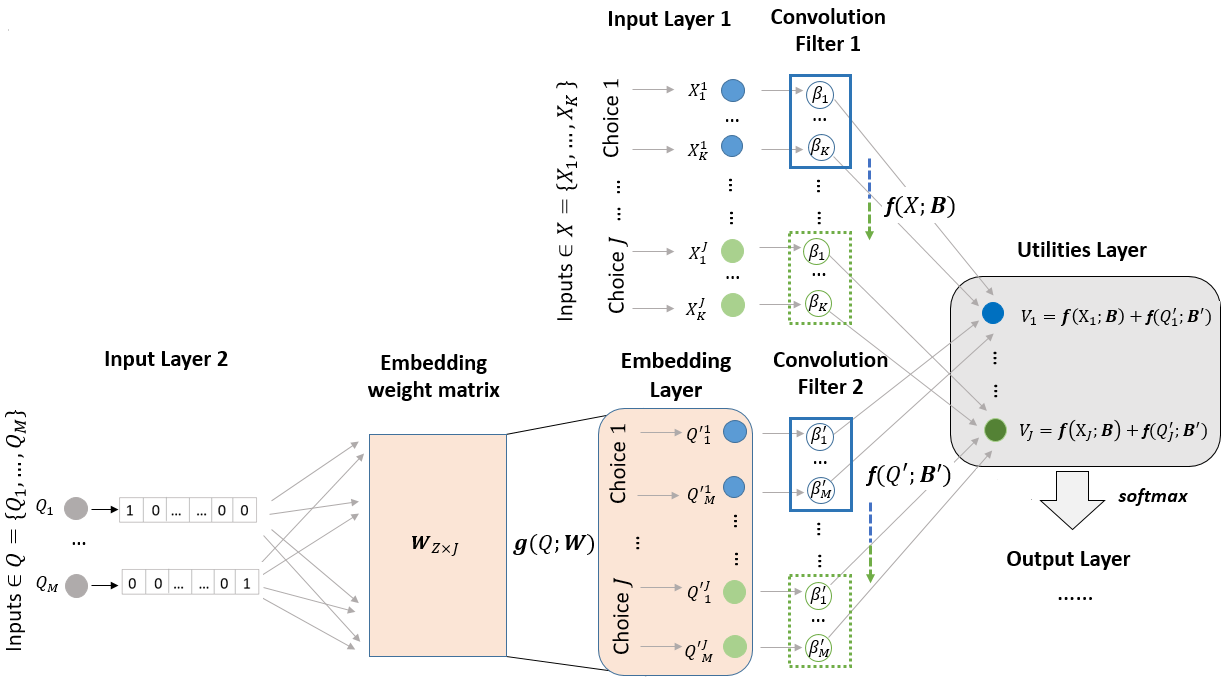}
    
    \end{tikzpicture}
    
    \caption{\textit{E-MNL model architecture.}}
\label{fig:mnl_emb}
\end{sidewaysfigure}

We redefine  $X=\{ X_{1}, X_{2}, ..., X_{K}\}$  to be a set of continuous explanatory variables and we consider $Q=\{Q_{1}, Q_{2}, ..., Q_{M}\}$  to be a set of categorical explanatory variables. Both can represent the observed attributes of the choice alternatives and the individual’s socio-demographic characteristics.
\newline Thus, the random utility $U$  that individual $n$ associates with alternative $i$ are given by:

\begin{equation}\label{eq5}U_{i,n}\left(X_{i,n}, Q_{i,n}\right)= V_{i,n}\left(X_{i,n}, Q_{i,n}\right) + \epsilon\end{equation}  
\newline
\newline We propose a ANN-based DCM model,  referred to as the Embeddings Multinomial Logit (E-MNL), such that the systematic part of the utility $V_{i,n}$ is expressed as:
\begin{equation}\label{eq6}V_{i,n}\left(X_{n}, Q_{n}\right)= \textit{\textbf{f}}_{i,1}(X_{n};\textbf{B}) + \textit{\textbf{f}}_{i,2}\left(\textit{\textbf{g}}_{i}(Q_{n};\mathbf{W}_{i});\textbf{B}')\right)\end{equation}
\newline\newline
The function $\textit{\textbf{g}}$ is defined as the embedding function, that maps each one-hot encoded input $Q_{n}$ to a latent 1-dimensional representation $Q'_{n}$ based on a set of trainable alternative-specific weights $\textbf{W}_{i}$, such that: $Q'_{i,n}=\textit{\textbf{g}}_{i}(Q_{n};\mathbf{W}_{i})$. 
\newline\newline The function $\textit{\textbf{f}}$ is defined as a linear function such that the trainable preference parameters of the model $\textbf{B}$ and $\textbf{B}'$ are a linear combination of the explanatory variables (or input features) $X_{i,n}$ and $Q'_{i,n}$ in $\textit{\textbf{f}}_{1}$ and $\textit{\textbf{f}}_{2}$ respectively, such that: $\textit{\textbf{f}}_{i,1}(X_{n};\mathbf{B})=\mathbf{B}X_{i,n}$ and  $ \textit{\textbf{f}}_{i,2}(\textit{\textbf{g}}_{i}(Q_{n};\mathbf{W}_{i});\mathbf{B'})=\mathbf{B'}Q'_{i,n}$.
\newline\newline The architecture of E-MNL is shown in Figure \ref{fig:mnl_emb}. We can observe that the network consists of two input layers receiving two separate inputs: $X$ and $Q$. In the first case the continuous inputs $X$ are directly connected to the first convolution filter and a set of trainable weights $\textbf{B}$. In the second case, the one-hot encoded inputs $Q$ are projected to the embedding layer that maps each input to a unique vector of dimensionality  $D=J$. Then, the new alternative-specific representations $Q'$ to are connected to the second convolution filter and a set of trainable weights $\textbf{B}'$. 
\newline\newline
The output of both convolution filters is combined to represent the systematic utilities of the model at the Utilities layer. We note that the size of the embedding matrix $\textbf{W}$ is $Z \times J$, where J is the number of choices in the choice set $C$ and $Z$ is the total number of unique categories across all the explanatory categorical variables $Q$, such that: $Z=\sum_{m=1}^{M}|Q_{m}|$. Lastly, for establishing that the signs of the embedding dimensions across $J$ are interpretable the preference parameters $\textbf{B}'$  are constrained to be positive.


Given the values of the model parameters $B$, $B'$ and $W$, and the input features $X_{n}$ and $Q_{n}$, the likelihood of selecting the choice alternative $i$ for individual $n$, is expressed as:
\begin{large}
\begin{equation}\label{eq7}P_{n}(i)=\cfrac{e^{\textit{\textbf{f}}_{i,1}(X_{n};\textbf{B}) + \textit{\textbf{f}}_{i,2}(\textit{\textbf{g}}_{i}(Q_{n};\textbf{W}_{i});\textbf{B}')}}{\sum_{j \in{C}_{n}}{e^{\textit{\textbf{f}}_{j,1}(X_{n};\textbf{B}) + \textit{\textbf{f}}_{j,2}(\textit{\textbf{g}}_{j}(Q_{n};\textbf{W}_{j});\textbf{B}')}}}\end{equation}
\end{large}
\newline
\newline
Although embedding representations of categorical variables for solving predictive tasks related to travel decision-making have been suggested before, our approach is novel in several respects summarized below.

\subsubsection{Interpretable embeddings dimensions}
First, our model produces embeddings representations of dimensionality $D$ that is equal to the number of alternatives $J$ in the choice set $C$, such that the embedding representation 
$Q'_{m}= [Q'_{m,1},Q'_{m,2}, ..., Q'_{m,J}]$, where $D=|Q'_{m}|= J $. Thus $D$ is fixed and predetermined by the number of choice alternatives and is not subject to tuning as it is usually the case for embeddings-based models, that treat $D$ as another hyperparameter of the model to optimize. More importantly, by formally associating each embedding dimension with one alternative, we yield embeddings that are interpretable such that the value of the embedding along to the j-th dimension represents the relevance of the encoded category to the j-th choice alternative. This approach allows us in practice to convert the categorical variables to continuous, as each of them has one continuous value for each alternative, rendering representations similar to e.g. "travel cost" or "travel time" that have differing values among different mode alternatives. This further allows them to share a single coefficient across the alternatives in the MNL specification \footnote{We should note that sharing a single coefficient across the alternatives is not a binding requirement for using the suggested model. The model specification can be modified to include alternative specific coefficients in the utilities by considering the corresponding alternative-specific value in each choice and 0 otherwise.}. Lastly, by restricting the coefficients associated to $Q'$ to be positive, we enhance the interpretability of the embedding values, such that the sign of an embedding value along the j-th dimension indicates the direction of the relevance/association between the encoded category and the j-th choice alternative, positive or negative.

\subsubsection{Joint embedding architecture}
In contrast to Pereira \cite{camara}  that uses separate embedding layers for each categorical variable considered, we use a single embedding layer where all the categories are projected to, regardless of the variable they originate from. This joint architecture produces categorical embeddings that are comparable both within and across variables, allowing us to: (i) compare the magnitudes of the MNL's coefficients across the categorical variables considered. (ii) identify meaningful clusters of categories in the embedding space that are related to the study behavior in a similar manner.


\subsection{Extended Model: Interpretable Embeddings MNL with Representation Learning term (EL-MNL) }

Lastly, we present the Embeddings Learning Multinomial Logit (EL-MNL)  that aims to combine the E-MNL model suggested in the previous section with the Learning Multinomial Logit (L-MNL) model suggested by Sifringer et. al in \cite{sin}. 

EL-MNL extends E-MNL model by increasing the dimensionality $D$ of the embedding matrix $W$ from $J$ to $J+S$ such that the extra dimensions $S$ are used as input to an additional densely connected hidden layer $L$ with $K$ hidden nodes $q_{1}, ... q_{K}$. $L$ produces an output a $J$-dimensional vector $r$, corresponding to a single term $r_{i}$ per utility function.

As suggested in \cite{sin}, $r_{i}$, referred to as the "representation learning term", is added to each of the utilities aiming to mitigate problems of endogeneity and correct for underfit due to undetected misspecification of the disturbance term, enchancing overall the predictive performance compared to a simple MNL and previously suggested ANN-based MNL model architectures.

\begin{sidewaysfigure}
\vspace*{7in}
    \begin{tikzpicture}\includegraphics [scale=1.5]{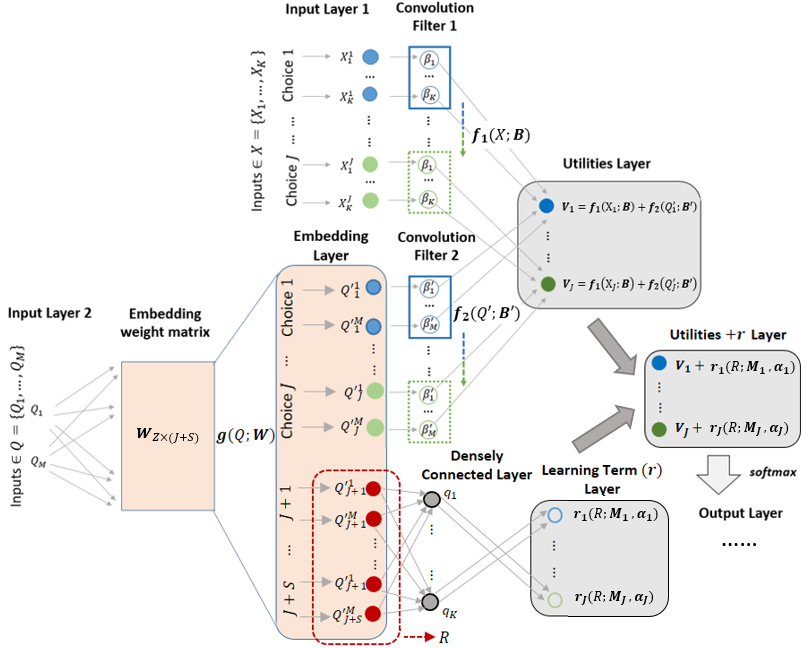}
    \end{tikzpicture}
    
    \caption{\textit{EL-MNL model architecture.}}
    \label{fig:el_mnl}
\end{sidewaysfigure}

In contrast to the L-MNL model architecture where the dense layer $L$ directly receives the raw data $Q$ as input without any intermediate transformation, in EL-MNL the dense layer $L$ receives as input \textit{a subset} of the output $Q'$ of the embedding layer. Specifically we increase number of embedding dimensions to be $D=J+S$, where $J$ is the number of choice alternatives (same as in E-MNL) and $S$ the number of additional embedding dimensions of $Q'$ that are intended to be used as an input to $L$. We will demonstrate that this extended model can be used to enhance predictability compared to both E-MNL and L-MNL, while maintaining interpretability for \textit{all} the input variables ($X$ and $Q$). 
\newline\newline
We thus propose an ANN-based DCM, EL-MNL, such that the systematic part of the utilities $V_{i,n}$ in (\ref{eq5}) is defined/expressed as:
\begin{equation}\label{eq8}\begin{aligned}V_{i,n}\left(X_{n}, Q_{n}\right)= \textit{\textbf{f}}_{i,1}\left(X_{n};\textbf{B}\right) +  \textit{\textbf{f}}_{i,2}\left(\textit{\textbf{g}}_{i}(Q_{n};\mathbf{W}_{i});\textbf{B}'\right) + \\ \textit{\textbf{r}}_{i}\left(\left\{\textit{\textbf{g}}_{J+1}(Q_{n};\mathbf{W}_{J+1}),...,\textit{\textbf{g}}_{J+S}(Q_{n};\mathbf{W}_{J+S})\right\};\textbf{M}_{i},\boldsymbol{\alpha}_{i}\right)\end{aligned}\end{equation}
The functions $\textit{\textbf{g}}$, and $\textit{\textbf{f}}$, as well as the trainable weights $\textbf{B},\textbf{B}', \textbf{W}$ are the same as defined in (\ref{eq6}) for E-MNL. The size of the embedding matrix $\textbf{W}$ is $Z \times D$, such that:
\newline\newline $Q'_{d,n}=\textit{\textbf{g}}_{d}(Q_{n};\mathbf{W}_{d})$ $\forall{d}\in{D}$ $(=J+S)$.
\newline\newline The function $\textit{\textbf{r}}$ represents the operations performed in the densely connected layer $L$ that receives $\left\{Q'_{J+1,n},...,Q'_{J+S,n}\right\}$ as an input.
For brevity and convenience we define the input to function $\textit{\textbf{r}}$ as $R_{n}=\left\{Q'_{J,n},...,Q'_{J+S,n}\right\}$
\newline\newline Thus (\ref{eq8}) can be rewritten as:

\begin{equation}\label{eq9}V_{i,n}\left(X_{n}, Q_{n}\right)= \textit{\textbf{f}}_{i,1}\left(X_{n};\textbf{B}\right) +  \textit{\textbf{f}}_{i,2}\left(Q'_{i,n};\textbf{B}'\right) + \textit{\textbf{r}}_{i}\left(R_{n};\textbf{M}_{i},\boldsymbol{\alpha}_{i}\right)\end{equation}
The function $\textit{\textbf{r}}$ uses a combination of alternative-specific trainable parameters $\textbf{M}_{i}$ and a bias term $\boldsymbol{\alpha}_{i}$, and the rectified linear unit (ReLU) activation function, such that: 
\begin{equation}\label{eq_r}\textit{\textbf{r}}_{i}\left(R_{n};\textbf{M}_{i},\boldsymbol{\alpha}_{i}\right)= \sum_{m \in \textbf{M}_{i}} k \left(\max(0, R_{n})\right) + \boldsymbol{\alpha}_{i}\end{equation}
\newline The architecture of EL-MNL is shown in Figure\ref{fig:el_mnl}.

Sifringer et.al in \cite{sin} have shown that an important constraint for obtaining interpretable parameters for $\textit{\textbf{f}}$, is that the set of features that enters the function $\textit{\textbf{r}}$, should not overlap to the ones that enter $\textit{\textbf{f}}$, which represents the interpretable part of the systematic utility. This constraint is met by our model formulation as shown in (\ref{eq8}) and (\ref{eq9}).
\section{Experiments}
Our experiments aim to compare E-MNL and the extended EL-MNL against benchmark models. The comparison takes place in four respects: predictive performance on unseen data, model complexity, interpretability, and model transparency. For assessing the first two we use  predictive performance log-likelihood (LL) on the test set as well as Akaike information criterion (AIC) that allows us to compare the models' expected performance while accounting for their complexity (total number of model parameters). The LL on the training set is also presented seperately to detect cases of potential overfitting. Interpretability concerns the extent to which a model is accessible to human understanding which, in our case, translates to the number of parameters that participate in the utility functions and the generated embedding values - since their dimensions are meaningful. Lastly, as a measure for each model's transparency, we use the ratio between the number of interpretable divided by the total number of parameters used in the model. Thus, in our case, model transparency refers to the extent to which the model parameter values are explainable and easily comprehended by a human analyst.

For our experiments we use two real-world datasets: swissmetro dataset and TU dataset. The first will be used to compare our models against the L-MNL model suggested by Sifringer et.al \cite{sin}, as well as a simple MNL using dummy encoding for the categorical explanatory variables. The Danish Travel Survey (Transportvaneundersøgelsen, TU) dataset is used as a case study to illustrate how E-MNL can improve the predictability while retaining full interpretability compared to the dummy encoding model for a simpler case of a binary forecasting choice problem. To establish the consistency of our results we use multiple subsets of the TU dataset sequenced over time. 

For both experiments, all the models were trained for 500 epochs each with 50 steps per epoch, using an Adam optimizer \cite{adam} with default settings and clipnorm of 50. Both in E-MNL and EL-MNL a regularization layer of 20\% dropout rate was used after the embeddings layer to prevent overfitting. We also note that, given the stochasticity of the networks' optimizer, we report average results over 30 training runs together with their standard deviation (std.).





\subsection{Multinomial case: Swissmetro Dataset - Predicting Mode Choice}

In this experiment we use swissmetro dataset \cite{swiss}, an openly available real-world dataset for mode choice offering three alternatives: Train, Swissmetro and Car. It consists of survey data collected on trains in Switzerland, during March 1998. The explanatory variables of the dataset used in the current paper are described in Table \ref{tab:variables}. 

\begin{table}[H]
\begin{center}
\begin{tabular}{ c c c}
\textbf{Variable} & \textbf{Description} &  \textbf{Type}\\ 
\hline
\hline
 PURPOSE  &  Trip purpose (business, leisure etc.) & Categorical\\ 
 FIRST  &    First class traveler & Binary\\
 TICKET & ticket type (one-way, half-day, etc.) & Categorical\\
 WHO & Who pays (self, employer etc.) & Categorical\\
MALE & Traveler’s gender & Binary\\
INCOME & Traveler's income level per year & Ordinal\\
LUGGAGE & Traveler's luggage pcs & Ordinal\\
AGE  & Traveler's age class & Ordinal\\ 
SEATS & Airline Seat configuration in the Swissmetro & Binary\\
GA & Annual season ticket & Binary \\
ORIGIN & Travel origin corresponding to a canton & Categorical\\
DEST  & Travel destination corresponding to a canton & Categorical\\
TT & Door-to-door travel time in minutes, scaled by 1/100 &  Continuous\\
TC & Travel cost in CHF, scaled by 1/100 &Continous\\
HEADWAY & Transportation headway in minutes, scaled by 1/100 &Continous\\
\hline
\hline
\end{tabular}
\end{center}

\caption{\textit{Variables in the Swissmetro dataset used in the current study.}}
\label{tab:variables}
\end{table}
\subsubsection{Data and Models specification}
The original dataset contains 10,728 observations. For comparison purposes we use the same initial dataset and data partition into a training a test set used in \cite{sin}. The observations for which the three  alternatives were not available, as well as the ones for which the information
regarding the chosen alternative was missing were discarded, resulting in a dataset of 9,036 observations. This dataset was further split into a training set and a test set of 7,234 and 1,802 observations respectively. 

Regarding the input feature sets for E-MNL and EL-MNL, we considered that $X$ contains the continuous variables presented in Table \ref{tab:variables} plus the intercept terms, such that $X=$ \{ASC$_{Car}$, ASC$_{SM}$, TT, TC, HEADWAY\}, and all the remaining 12 variables constitute the feature set $Q$ that are projected to the the embedding layer. We note that the number of unique categories across all the variables in $Q$ is $Z= 81$.

\subsubsection{Training the models}

We trained 30 $\times$ E-MNL models and 2 $\times$ 30 $\times$ EL-MNL models (we considered 2 different hyperparameter configurations) using Keras python deep learning library \cite{keras}. For E-MNL the number of embedding dimensions $D$ is defined by the number of alternatives in the choice set and thus $D=3$. For the hyperparameters of the extended model EL-MNL we considered the number of hidden nodes in the additional layer $L$, to be $K=15$ and the number of extra embedding dimensions $S$ to be 1 and 2, resulting in embeddings dimensionality $D$ of 4 and 5 respectively. 

\subsubsection{Results and Comparison to Benchmark Models}

As benchmarks for comparison purposes in this experiment we consider the following models:
\newline
\newline(i) MNL$_{orig}$ using the original utility specification from Bierlaire et al. \cite{swiss}. 
\newline
\newline(ii) MNL$_{dum}$ with inputs $X=$ \{ASC$_{Car}$,ASC$_{SM}$,TT, TC, HEADWAY\} and $Q= $\{PURPOSE, FIRST, TICKET, WHO, MALE, INCOME, LUGGAGE, AGE, SEATS, GA, ORIGIN, DEST\}. The variables in $Q$ are dummified. We consider homogeneous and alternative-specific coefficients for the variables contained in $X$ and $Q$ respectively \footnote{We also note that in order to avoid singularities in the Hessian matrix the following dummy predictors were removed from the model specification: O\_GL, O\_LU, O\_SZ, D\_SZ, Other\_purp, FromLeisure.}.
\newline
\newline(iii) L-MNL$_{1}$ (as presented by Sifringer et. al. in \cite{sin}) with $K=100$, and inputs $X=$ \{TT, TC, HEADWAY\} and $Q= $\{PURPOSE, FIRST, TICKET, WHO, MALE, INCOME, LUGGAGE, AGE, SEATS, GA, ORIGIN, DEST\}.
\newline
\newline(iv) L-MNL$_{2}$ (as presented by Sifringer et. al. in \cite{sin}) with $K=100$, and inputs $X=$ \{TT, TC, HEADWAY, AGE, LUGGAGE, SEATS, GA\} and $Q= $\{PURPOSE, FIRST, TICKET, WHO, MALE, INCOME, ORIGIN, DEST\}.
\newline

In Table ~\ref{tab:results swissmetro}, we present the comparison between E-MNL, EL-MNL and all the benchmark models with respect to predictive log-likelihood (LL) on the test set as well as Akaike information criterion (AIC). The LL on the training set is also presented to detect cases of potential overfitting. Given the fact that we trained multiple models for E-MNL and EL-MNL, their average LL on the training and the test set and AIC are presented, together with their corresponding standard deviations. We note that the LL results reported for models (i), (iii) and (iv) are reprinted from \cite{sin}, while the ones for (ii) were estimated using Pylogit \cite{pylogit}.
\begin{table}[H]
\small
\begin{center}
\begin{tabular}{ c |c c| c c c| c}
\textbf{Suggested} & \textbf{$\overline{LL}_{train}$} &  \textbf{$\overline{LL}_{test}$} &
interpret.&total&ratio&$\overline{\mbox{AIC}}$\\
\textbf{Models}&(std.) &(std.)&params&params&&(std.)\\
\hline
\hline
E-MNL($D=$ 3)& $-4912.7$& $-1231.1$&$260$&$260$&1&$10345.5$\\
& $(3.0)$& $(1.1)$&&&&$(6.0)$\\
\hline
EL-MNL($D=$ 4, $K=$ 15)& $-4393.8$ &$ -1134.9$ &$260$&$584$&0.45&$9961.4$\\
&$(34.2)$& $(8.3)$&&&&$(69.3)$\\
\hline
EL-MNL($D=$ 5, $K=$ 15)&$ -4091.6 $& $\textbf{-1097.7}$&$260$&$845$&0.31&$\textbf{9876.4}$\\
&$(40.7)$&$(13.8)$&&&&$(80.5)$\\
\hline
\textbf{Benchmark}&&&&&\\
\textbf{Models}&\textbf{$LL_{train}$}&\textbf{$LL_{test}$}&&&&AIC\\
\hline
\hline
MNL$_{orig}$ &$-5764 $&$-1433$&$9$&$9$&1&$11546$\\

MNL$_{dum}$ &$-4903$ & $-1928$&$130$&$130$&1&$10066$\\
L-MNL$_{1}$($K=$ 100)&$-4511$ &$-1181$&$9$&$1202$&0.007&$ 11426$\\
L-MNL$_{2}$($K=$ 100)&$-3895$& $-1108$&$3$&$1606$&0.002&$11002$\\
\end{tabular}
\end{center}

\caption{\textit{Comparison of log-likelihood, AIC and number of parameters between the suggested and benchmark models on swissmetro dataset.}}
\label{tab:results swissmetro}
\end{table}
The AIC results show that all the suggested models have a better, more parsimonious fit compared to all the benchmark models, except MNL$_{dum}$, which however is a typical case of overfitting, as its expected predictive performance is not reflected in the test set results, which can be attributed to the high dimensionality of the feature space. In terms of predictive performance, we found that the EL-MNL that uses 5 embedding dimensions yields the best overall results across all the considered models. The E-MNL has a higher LL on the test set compared to the simple MNL models MNL$_{orig}$ and MNL$_{dum}$ by 15.6\% and 44.1\% respectively, while it shows a relatively lower predictive performance compared to the L-MNL models suggested in \cite{sin}. However, when we use the extended model architecture, the EL-MNL models with 4 and 5 number of embedding dimensions manage to outperform L-MNL$_{1}$ and L-MNL$_{2}$ respectively, while using approximately half the number of parameters as them. These results suggest that combining the embeddings encoding with inclusion of the representation learning component in the utility specification into a single model (EL-MNL) can lead to a model that performs better than the underlying models individually (with an increase by about 11.5\% , 7.3\% and 0.94\%  compared to E-MNL, L-MNL$_{1}$ and L-MNL$_{2}$ respectively), while limiting substantially the number of the required parameters in L-MNL$_{2}$ for achieving the same performance levels. 

Except for the advances in performance, the proposed embedding-based models offer richer, interpretable outputs for \textit{all the} 15 \textit{input explanatory variables}, while the meaningful network parameters in L-MNL model are \textit{limited to the utility coefficients for the inputs in $X$} (corresponding to 3 and 9 parameters in L-MNL$_{1}$ and  L-MNL$_{2}$ respectively). The interpretable parameters for our models consist of the utilities coefficients for the variables both in $X$ and $Q$, and the alternative-specific embedding values for the categorical variables in $Q$. The former allows us to quantify the unique contribution of each of the input variables to the models' predictions, and the latter provides us meaningful representations of the internal variable relationships that can offer further insights into understanding the observed behavior. 

Lastly, in terms of transparency, the only ANN-based model that is fully transparent, similarly to simple MNL models, is the E-MNL model. We can further observe that the L-MNL models suggested in \cite{sin} have a very low degree of transparency (0.002 and 0.007) compared to the proposed EL-MNL models that manage to preserve a relatively high transparency (0.31 and 0.45)  despite their augmented hidden layer architecture. 
Overall, the extended EL-MNL architecture comprises a compact and highly transparent model that manages to combine all the benefits of the different models presented that can be summarized  as follows: 
(i) utility coefficients for all the input explanatory variables, as in simple MNL models.
(ii) informationally rich representations in the form of alternative specific embedding values for all the input variables in Q, as in E-MNL. 
(iii) high predictive power, as in L-MNL. 

A final remark on Table 2 concerns the standard deviation (std) in the results of the proposed models over 30 training runs. We can observe that the E-MNL architecture renders a model that produces less volatile results compared to the ones of EL-MNL models. This indicates that the use of a hidden layer to derive the non-linear learning term $r$ is related to the model's ability to produce stable results over multiple training runs, which seems to decrease with model complexity. Although in the case of the E-MNL stability is not an issue, the higher std. values in EL-MNL's results, suggest that the model selection process would ideally require a validation set to be used for the optimization of the model's parameters\footnote{{We note that in the current experiment we avoided using a validation set, and followed the same dataset partition as in \cite{sin} in order to keep our results comparable.}}. 

In Table 3 we present the parameters estimates for E-MNL and EL-MNL models with the best predictive performance in the test set. We observe that the embeddings encoding models, not only yield a better fit and improved out-of-sample performance compared to the benchmark models, but also produce significant parameter estimates for the majority of the categorical variables in $Q$, as well as expected negative signs for TT, TC and HEADWAY which are consistent with theoretical expectations. In the best performing model, 3 variables (MALE, GA and SEATS) were identified as statistically insignificant predictors, while highly significant parameter estimates (at 0.001 Sig level) were obtained for all the remaining explanatory variables in $Q$ and $X$. It is worth-noting that the ORIGIN and DEST variables, which were omitted in the original model specification in \cite{swiss}, yield not only statistically significant but also the largest beta values across all the variables in $Q$, indicating that they contain categories that are strong predictors for the task at hand. The high cardinality and sparsity of these variables, however, poses challenges for including them in the model specification as dummies. The proposed embedding-based models offer a compact and elegant solution for not only including such variables in the modelling process but also quantifying their contribution in statistical terms.

\begin{table}[H]
\fontsize{9}{10}\selectfont
\begin{center}
\begin{tabular}{ c c c c c c c}

\textbf{\begin{small} Model\end{small}} & \textbf{\begin{small} Params\end{small}} &  \textbf{\begin{small}Estimates\end{small}}& \textbf{\begin{small}Std errors\end{small}} & \textbf{\begin{small}t-stat\end{small}} & \textbf{\begin{small}p-value\end{small}} & \textbf{\begin{small}Sig.\end{small}}\\
\hline
\hline
\\
\begin{small}E-MNL($D=$ 3)\end{small} &  ASC$_{Car}$ &0.252	&0.093	&2.722	&0.006&***\\
\begin{small}LL$_{train}$: $-4911.4$\end{small}&ASC$_{SM}$ &0.476	&0.079 &6.022	&0.000&***\\
\begin{small}LL$_{test}$:  $-1229.2$ \end{small}&TT &-1.536	&0.054	&-28.437 &0.000&***\\
&TC &-1.129	&0.043	&-26.099		&0.000&***\\
&HEAD$_{Train,SM}$ &-0.808	&0.122	&-6.647	&0.000&***\\
&PURPOSE &14.074	&1.052	&13.377	&0.000&***\\
&FIRST &2.059	&1.142	&1.802		&0.072&*\\
&TICKET &12.952	&0.676	&19.167	&0.000&***\\
&WHO &8.256	&1.392	&5.929		&0.000&***\\ 
&LUGGAGE &7.118	&2.428	&2.931	&0.003&***\\
&AGE &8.484	&0.894	&9.494	&0.000&***\\
&MALE &3.752 &2.078	&1.806	&0.071&*\\
&INCOME &7.417	&1.124	&6.599	&0.000&***\\
&GA &7.429	&1.479	&5.022		&0.000&***\\
&ORIGIN &15.325	&1.949	&7.862 &0.000&***\\
&DEST &15.425	&1.572	&9.814 &0.000&***\\
&SEATS &7.526	&1.614	&4.662	&0.000&***\\ 
\hline
\\
\begin{small}EL-MNL($D=$ 4, $K=$ 15)\end{small} &  ASC$_{Car}$ &0.084	&0.096	&0.879	&0.380& \\
\begin{small}LL$_{train}$: $-4330.0$\end{small}&ASC$_{SM}$ &-1.178	&0.082	&-14.430	&0.000&***\\
\begin{small}LL$_{test}$: $-1114.4$\end{small} &TT &-1.740	&0.056	&-31.260	&0.000&***\\
&TC &-1.583	&0.049	&-32.490	&0.000&***\\
&HEAD$_{Train,SM}$ &-0.856	&0.126	&-6.777 &0.000&***\\
&PURPOSE &14.436	&1.636	&8.826 &0.000&***\\
&FIRST &3.057	&0.888	&3.442 &0.001&***\\
&TICKET &14.423	&1.373	&10.503 &0.000&***\\
&WHO &8.909	&2.207	&4.037& 0.000&***\\ 
&LUGGAGE &6.622	&2.486	&2.663	&0.008&***\\
&AGE &7.516	&1.249	&6.017	&0.000&***\\
&MALE &1.940	&3.040	&0.638	&0.523\\
&INCOME &6.919	&1.380	&5.015	&0.000&***\\
&GA &5.768	&1.627	&3.546	&0.000&***\\
&ORIGIN &17.468	&2.148	&8.133 &0.000&***\\
&DEST &18.367	&1.645	&11.168	&0.000&***\\
&SEATS &5.367	&3.254	&1.649	&0.099&*\\
\hline
\\
\begin{small}EL-MNL($D=$ 5, $K=$ 15)\end{small} &  ASC$_{Car}$ &0.239	&0.097	&2.464	&0.014&** \\
\begin{small}LL$_{train}$: $\mathbf{-4029.4}$\end{small}&ASC$_{SM}$ &-1.429	&0.081	&-17.594	&0.000&***\\
\begin{small}LL$_{test}$: $\mathbf{-1072.5}$\end{small} &TT &-1.919	&0.055	&-34.876&0.000&***\\
&TC &-1.905	&0.053	&-35.665	&0.000&***\\
&HEAD$_{Train,SM}$ &-0.856	&0.126	&-6.767&0.000&***\\
&PURPOSE &10.806	&1.496	&7.226	&0.000&***\\
&FIRST &3.408	&0.859	&3.966&0.000&***\\
&TICKET &13.931	&1.956	&7.121&0.000&***\\
&WHO &7.242	&1.680	&4.311	& 0.000&***\\ 
&LUGGAGE &6.257	&1.918	&3.262&0.001&***\\
&AGE &7.519	&1.274	&5.901&0.000&***\\
&MALE &0.928	&3.811	&0.244	&0.808\\
&INCOME &6.325	&1.344	&4.707	&0.000&***\\
&GA &1.134	&11.602	&0.098	&0.922&\\
&ORIGIN &16.326	&2.195	&7.439	&0.000&***\\
&DEST &17.645	&1.718	&10.269	&0.000&***\\
&SEATS &2.697	&1.993	&1.353	&0.176&*\\
\hline
\hline
\end{tabular}
\end{center}
\caption{\textit{Log-likelihood and parameters estimates for E-MNL and EL-MNL with the best predictive performance in the test set. Significance level $(Sig): *** p < 0.001; ** p < 0.01; * p < 0.05$.}}
\label{tab:variab}
\end{table}
\subsection{Binary Case: TU Dataset - Predicting Car Ownership on a Household Level}\label{tu_experiment} 
For the purposes of predicting household car ownership,  we used a subset of the Danish National Travel Survey-Transportvaneundersøgelsen (TU) dataset  covering the period from 2014 to 2018\footnote{For survey description please visit: https://www.cta.man.dtu.dk/english/national-travel-survey/}. 
The variables used for training the models are presented in Table~\ref{tab:TU_variables}.
\begin{table}[H]
\small
\centering
\begin{tabular}{cccc}
\textbf{Discrete explanatory Variables ($Q$)} & \textbf{Categories or Intervals}\\
\hline
\hline
{\textit {HouseType}} & {detached, farm, flatblock} \\
(respondent's housing type) & {linked, other, studenthus} \\
\hline
\hline
{\textit {BinFamN}} & {[1, 2], (2, 3], (3, 4], (4, 13]} \\
(respondent's family size) -  \textit{discretized}&{} \\
\hline
\hline
{\textit {NuclFamType}} & {SingleM, SingleW, Couple,} \\
(respondent's nuclear family type) & {Single\&Children, Couple\&Children} \\
\hline
\hline
{\textit{HousehAccOwnOrRent}} & {cooperative, owner, rent}\\
{(respondent's type of house ownership)} & \\
\hline
\hline
{\textit{HomeAdrNUTS}} &  {WZealand, CPH, EJutland, Funen, SJutland,}\\
{(region of household address)}& {NZealand, EZealand, NJutland, GrCPH, WJutland}\\
\hline
\hline
{\textit{RespPrimOcc}} & {Pension, SelfEmpl, EarlyPension, Employee,} \\
{(respondent's primary occupation)}& {nonAgePension, Unempl, Student, HighSchool,}\\
& {PrimSchool,OtherOccupation}\\
\hline
\hline
{\textit{RespEduLevel}} & {1st-7th, 8th, 9th, 10th, MediumFurtherEdu,}\\
{(respondent's highest} & {LongFurthEdu, Upper2ndCertif, HigherCertif,}\\
{completed education)} & {Vocational, OtherSchool}\\
\hline
\hline
{\textit{IncNuclFamily}} & {[0, 250], (250, 400], (400, 550]}, \\
 {(household's total gross income} & {(550, 700], (700, 870]}\\
 {in thousand DKK per year)- \textit{discretized}}& \\
\hline
\hline
{\textit{HomeParkPoss}} & {e.g. on private lot, on street/road}\\
 {(parking conditions at home)} & {rarely/never space but free ...}\\
\hline 
\hline 
{\textit{WorkParkPoss}} & {e.g. always space free parking,}\\
 {(parking conditions at } & {normally space payment required ...}\\
 {at place of occupation)}&\\
\hline 
\hline 
{\textbf{Continuous explanatory Variables ($X$)}} & {\textbf{Values range}}\\
\hline
\hline
{\textit{HomeDistNearestStation}} &  \\
{(distance from respondent's residence} &{[0.0, 53.4]}\\
{to nearest station in km)} &\\
\hline 
\hline 
{\textit{GISdistHW}} &  \\
{(distance from respondent's residence} &{[0.0, 383.89]}\\
{to work in km)} &\\
\hline 
\hline 
{\textbf{Target variable}} & {\textbf{Values}}\\
\hline 
\hline
{\textit{Car Ownership}} & {No Car (0), Car (1)}\\
{\textit{on a household level} }&\\
\end{tabular}

\caption{\textit{The TU variables used together with their corresponding values, categories or intervals.  (For the full list of categories for the variables \textit{HomeParkPoss} and \textit{WorkParkPoss} see please visit the documentation of the TU dataset).}}
\label{tab:TU_variables}
\end{table}
\subsubsection{Data and Models specification}
After excluding the observations for which some variables were not available we ended up with a total of 14,550 observations. We divided the data into 5 subsets of yearly periods. All the models in this experiment were trained on the data for each of the years from 2014 to 2017 and tested on the data for the following year in each case.
For the input feature sets for models E-MNL and EL-MNL, we considered that $X$ contains the continuous variables presented in Table~\ref{tab:TU_variables} plus the intercept term, such that $X=$ \{ASC$_{Car}$, \textit{HomeDistNearestStation}, \textit{GISdistHW}\}, and the remaining 10  discrete variables constitute the feature set $Q$ with $Z= 79$ the number of unique categories across all the variables in $Q$ .
\subsubsection{Training the models}
We trained 30 $\times$ E-MNL models and 30 $\times$ EL-MNL models for each yearly subset of the data (240 models in total). Since the target variable is a binary choice of whether a household owns a car or not, we considered $D=2$ as the number of embedding dimensions $D$ for the input categories in $Q$, i.e. one embedding dimension associated with car ownership ($Q'_{Car}$) and one associated with no car ownership ($Q'_{NoCar}$).

However, given that we use a simple binary logit model with no alternative-specific variables, the number of trainable parameters in the considered models can be restricted by imposing $Q'_{Car,m}=-Q'_{NoCar,m}$ for every input category $m$ in $Q$, which is equivalent to imposing equal and opposite coefficients estimates between the binary alternatives.
Additionally, for the extended model EL-MNL, we consider the simplest configuration\footnote{Although we experimented with different values of of $K$ and $S$, we noticed that increasing the values of the model's hyperparameters results in degrading the model's performance.} using $K=1$ as the number of hidden nodes in the additional layer $L$, and  $S=1$ as the number of extra embedding dimensions, such that $D=3$.

\subsubsection{Results and Comparison to dummy-encoding MNL}

As baseline for comparison purposes we consider a simple binary logit model, referred to as MNL$_{dum}$ with continuous inputs $X=$ \{ASC$_{Car}$, \textit{HomeDistNearestStation}, \textit{GISdistHW}\}, and the remaining 10 discrete variables $Q$ represented as dummies. In Table~\ref{tab: TU results} we present and compare the LL and AIC results 
between E-MNL, EL-MNL and the baseline model, while in Table~\ref{tab: TU params} the interpretable and total number of models' trainable parameters, as well as their ratio is presented. 

We can observe that in terms of predictive performance (LL$_{test}$) the suggested models perform consistently better compared to the baseline model  for all the datasets considered. However, the range of the improvement over the baseline varies from a fractional 0.49\% in Experiment 2, to a substantial 45.53\% and 20.27\% gain in Experiments 1  and 4 respectively. In both these experiments we can observe typical cases of overfitting for the baseline dummy-encoding model, since the LL observed on the training data drops dramatically when the model is evaluated on the test set. Such models are not suitable for forecasting applications, indicating that the model selection by means of the AIC, that is lower for the baseline model in the 3 out of 4 Experiments, may be misleading. The problem of overfitting is not present in the suggested models that use embedding encoding for the discrete variables. This can be attributed to the fact that the embeddings encoding significantly reduces the dimensionality of the feature space and produces dense, meaningful representations, thus the embeddings-based models are less susceptible to random errors and noise in the data.

When comparing the 2 proposed models, we can observe that they achieve almost the same predictive performance levels. E-MNL exhibits overall slightly better LL on the test data than the extended EL-MNL model, ranging between 0 (in Experiment 2) to 0.98\% (in Experiment 4). In addition to not contributing to performance enhancement, EL-MNL has almost double the number of trainable parameters compared to E-MNL, and its transparency (number interpretable to total parameters ratio) drops from 1 (fully transparent model) to 0.54. These results and considerations suggest that the use of the extended architecture model EL-MNL may be redundant depending on the complexity of the dataset and the task at hand, and thus its use should be considered as complementary to that of E-MNL and should be subject to investigation. 
\begin{table}
\footnotesize
  \begin{tabular}{l|lll|lll}
  \multicolumn{1}{c}{ \textbf{Models}} & 
  \multicolumn{2}{c}{ \textbf{Experiment 1}}&\multicolumn{4}{c}{ \textbf{Experiment 2}}\\
 &Train-year 2014& &Test-year 2015 & Train-year 2015& &Test-year 2016\\
  &(n obs= 2,619)&&(n obs= 2,373) &(n obs= 2,373)&& (n obs= 3,033)\\
  &&&&&&\\
  & \textbf{$\overline{LL}_{train}$} & $\overline{\mbox{AIC}}$ & \textbf{$\overline{LL}_{test}$}& \textbf{$\overline{LL}_{train}$} & $\overline{\mbox{AIC}}$ & \textbf{$\overline{LL}_{test}$} \\
  &(std.)&(std.)&(std.) &(std.)&(std.)&(std.)\\
  E-MNL($D=$ 2)& -642.3 & 1468.6 & \textbf{-601.1}&-537.4 & \textbf{1258.8}& \textbf{-697.6}\\
    &(1.5)&(3.0)&(2.3)&(0.6)&(1.3)&(2.2)\\
 &&&&&&\\
  & \textbf{$\overline{LL}_{train}$} & $\overline{\mbox{AIC}}$ & \textbf{$\overline{LL}_{test}$}& \textbf{$\overline{LL}_{train}$} & $\overline{\mbox{AIC}}$ & \textbf{$\overline{LL}_{test}$} \\
  &(std.)&(std.)&(std.) &(std.)&(std.)&(std.)\\ 
  EL-MNL($D=$ 3, &\textbf{-636.8}  & 1645.6& -602.4& \textbf{-534.2} &1440.4 & \textbf{-697.6}\\
 \hspace{1.5cm}$K=$ 1) &(12.4)&(24.9)&(6.5)&(9.0)&(17.9)&(4.4)\\
 &&&&&&\\
  & \textbf{${LL}_{train}$} & AIC & \textbf{${LL}_{test}$} &
    \textbf{${LL}_{train}$} & AIC& \textbf{${LL}_{test}$}\\

   MNL$_{dum}$& -635.3 &  \textbf{1414.5}& -955.4&-558.0&1260.0& -701.0 \\
 &&&&&&\\
 &&&&&&\\
 \multicolumn{1}{c}{ \textbf{Models}} & 
  \multicolumn{2}{c}{ \textbf{Experiment 3}}&\multicolumn{4}{c}{ \textbf{Experiment 4}}\\
 &Train-year 2016& &Test-year 2017& Train-year 2017& &Test-year 2018\\
  &(n obs= 3,033) &&(n obs= 3,231) &(n obs= 3,231)&& (n obs= 3,294)\\
  &&&&&&\\
  & \textbf{$\overline{LL}_{train}$} & $\overline{\mbox{AIC}}$ & \textbf{$\overline{LL}_{test}$}& \textbf{$\overline{LL}_{train}$} & $\overline{\mbox{AIC}}$ & \textbf{$\overline{LL}_{test}$} \\
  &(std.)&(std.)&(std.) &(std.)&(std.)&(std.)\\
 E-MNL($D=$ 2)& -618.6 & 1421.2&  \textbf{-690.1} &-629.4 & 1442.8 &  \textbf{-854.3}\\
    &(0.6)&(1.1)&(1.4)&(0.5)&(1.1)&(2.4)\\
 &&&&&&\\
  & \textbf{$\overline{LL}_{train}$} & $\overline{\mbox{AIC}}$ & \textbf{$\overline{LL}_{test}$}& \textbf{$\overline{LL}_{train}$} & $\overline{\mbox{AIC}}$ & \textbf{$\overline{LL}_{test}$} \\
  &(std.)&(std.)&(std.) &(std.)&(std.)&(std.)\\ 
  EL-MNL($D=$ 3, &\textbf{-607.0}  & 1586.0 & -693.9& \textbf{-615.8} & 1603.6& -862.7\\
 \hspace{1.5cm}$K=$ 1) &(9.2)&(18.5)&(7.0)&(16.1)&(32.1)&(14.8)\\
 &&&&&&\\
  & \textbf{${LL}_{train}$} & AIC & \textbf{${LL}_{test}$} &
    \textbf{${LL}_{train}$} & AIC& \textbf{${LL}_{test}$}\\

    MNL$_{dum}$& -614.6&  \textbf{1373.2}& -703.6 &-624.0& \textbf{1392.1} &-1047.0\\

  \end{tabular}
\caption{\textit{Comparison of log-likelihood and AIC on TU datase between the suggested models and a simple MNL that uses dummy encoding for the variables in $Q$.}}
\label{tab: TU results}
\end{table}
\begin{table}
\small
  \begin{tabular}{l|l|l|l}
  \textbf{Models}& \textbf{Interpretable} & \textbf{Total} & \textbf{Ratio}\\
  &\textbf{params}&\textbf{params}&\\
  \hline
  E-MNL($D=$ 2) & 92 & 92 & 1\\
  EL-MNL($D=$ 3,$K=$ 1) &92 & 171 & 0.54\\
  MNL$_{dum}$& 71 & 71 & 1\\
  \end{tabular}
\caption{\textit{Comparison of number of total parameters, interpretable parameters and their ration (indicating model transparency) between the suggested models and a simple MNL that uses dummy encoding for the discrete expanatory variables in TU dataset.}}
\label{tab: TU params}

\end{table}
\newpage\section{{Embeddings: Interpretability}}
\begin{figure}
    \subfloat[]{\label{sublable1}\includegraphics[height = 3.3 in]{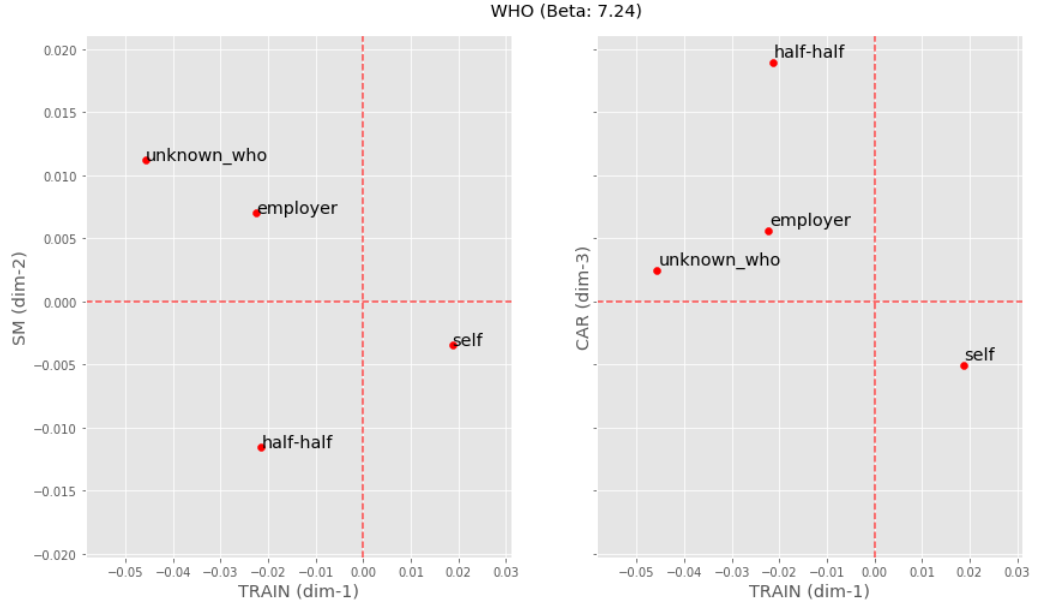}} \\
    \subfloat[]{\label{sublable2}\includegraphics[height = 5.6 in]{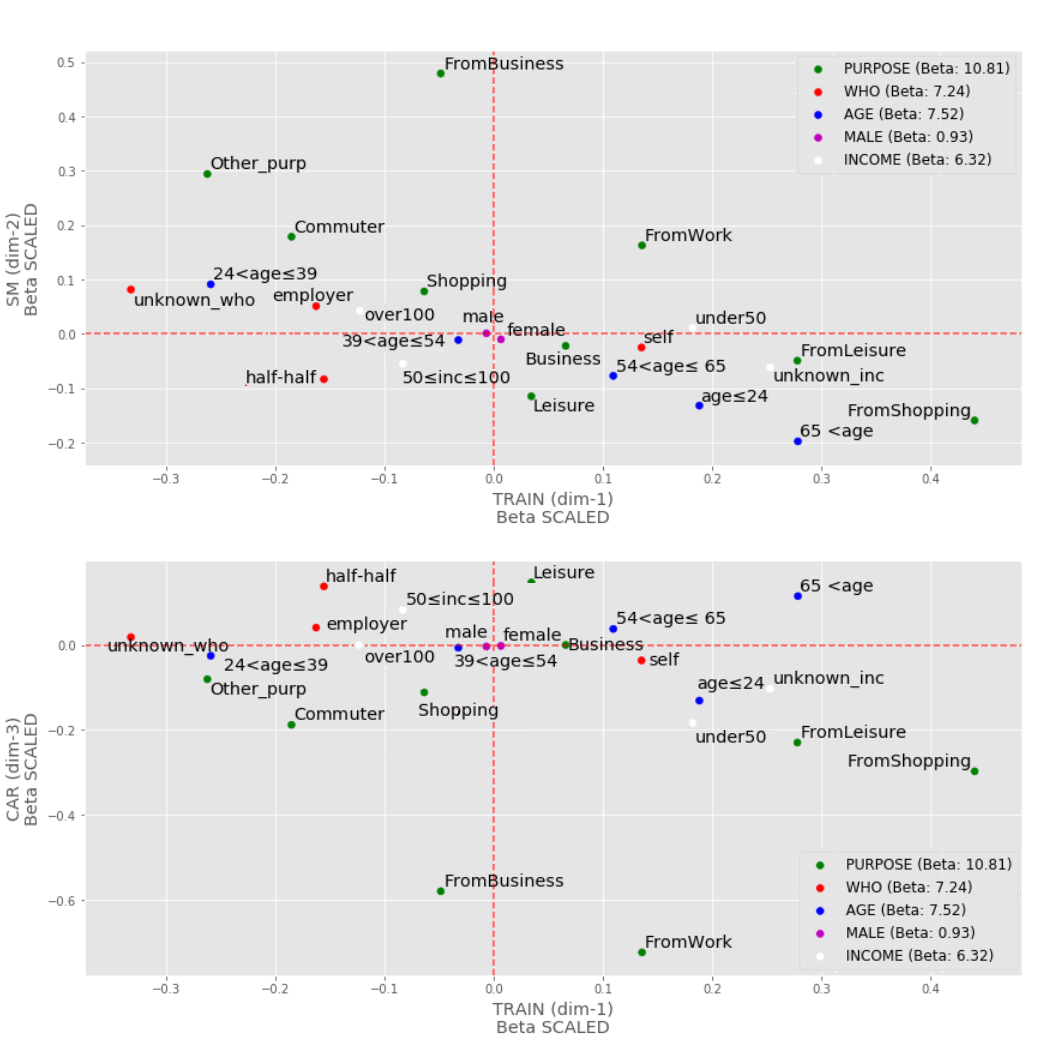}}
    \caption{\textit{Comparison of the categorical embeddings \textit{within} and \textit{across} variables.\newline \textbf{(A)} Visualization of the embeddings dimensions for WHO \textit{(pays)} variable.\newline \textbf{(B)} Common space visualization of the embeddings dimensions for multiple variables in \textit{swissmetro} scaled by their respective beta values.}}
    \label{visualization}
\end{figure}

In order to further understand embeddings, and interpret their meaning from a choice modeling perspective, we visualize the swissmetro dataset categorical variables in Figure \ref{visualization} according to their representation in the embeddings space (it is 3-dimensional because there exists 3 mode choice alternatives). In Figure \ref{sublable1} we plot the embedding values for the categories corresponding to the variable WHO (i.e. who pays the travel price), while in  Figure \ref{sublable2} the ones for the categories corresponding to several other variables\footnote{ For readability and to avoid cluttered figures, we only included a subset of categorical variables of the swissmetro dataset in the visualizations presented in this section.}. The embedding values presented in these figures are scaled by their respective beta values for comparison purposes.
\newline For interpreting such visualizations we need to focus both on the sign and value each category receives along each dimension of the embedding space, as well as the relative distances between every pair of points. Higher and positive values along each alternative-specific axis show a higher positive effect of the visualised categories on the corresponding choice alternative, while closer proximity between categories in the space indicates that they have a similar overall effect from the perspective of mode choice.

Figure \ref{sublable1} allows us to directly compare how the different categories within the same variable, affect mode choice and how they are related to each other. We can observe for example that the travellers who pay the whole travel price on their own (\textit{self}) prefer to use train, which is the cheapest alternative. However, when the travel price is wholly or partially covered by their employer, they tend to avoid train as the negative sign along its corresponding axis indicates, and resort to more expensive but also more convenient and faster mode options. In specific, when the employer covers half of the price (\textit{half-half}), the individuals tend to prefer travelling by car, and avoid using the swissmetro, while in the case that the travel cost is fully covered by their employer), they tend to choose either swissmetro or car, showing a slightly higher preference for the former. 

In Figure \ref{sublable2} the embedding values for the categories corresponding to 5 swissmetro variables (WHO, PURPOSE, AGE, MALE, and INCOME) are plotted together in  a common space. Such a visualization allows us to make direct comparisons regarding the contribution of the input variables to the travel mode choice predictions, as well as how categories across different variables affect the observed behavior. Overall, we can notice that, from the plotted variables, the ones that seem to be more relevant for choosing the new mode of transport (swissmetro) are PURPOSE and AGE (green and blue dots in the plot), since they contain categories that have the highest and lowest embedding values along the corresponding SM dimension. Among these categories, we can identify the ones that are the most and the least relevant for choosing swissmetro: returning from a business-related trip (\textit{FromBusiness}) and being above 65 years old (\textit{65<age}). \textit{FromBusiness} category seems to have a negative association with choosing the other 2 modes, a lower one for train and a much more negative one for car. The people in the \textit{65<age} category on the other hand, would avoid using the swissmetro in favor of either train or (less often) car.  

\begin{figure}
\centering
\captionsetup{justification=centering}

 \includegraphics[scale=1.1]{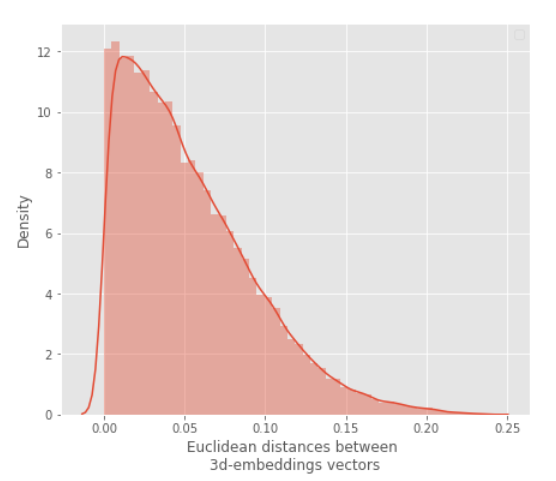}
\caption{\textit{Pairwise distribution of euclidean distances between the (beta scaled) embedding vectors for the categories of the variables PURPOSE, WHO, AGE, MALE, and INCOME.}}
\label{fig:eucl}
\end{figure}

Another way of interpreting such visualizations needs to focus on similarity between sets of categories. If two categories are very close in the embedding space, it means that, from the perspective of mode choice, they have a similar effect. The proximity between two categories in the space can be measured through a distance metric between their corresponding embedding vectors, such as euclidean distance. For diagnostic and analysis purposes, it is often useful to plot the distance distribution for the embeddings matrix. In Figure~\ref{fig:eucl} we have plotted the pairwise distance distribution between the embedding vectors for all the categories shown in Figure ~\ref{sublable2}. If this distribution is concentrated around the same value it indicates that the encoded categories fall in roughly the same distance from each other, i.e. they are equidistant, thus using embeddings encoding will not add much value in comparison to dummy encoding. A particular case is when there is high concentration around zero. This means that there are pairs or clusters of  categories that are associated with the target variable in a similar manner, suggesting that at least some of the categories can just be ommitted or grouped together into a single one (e.g. a single dummy variable).

\begin{figure}
\centering

\includegraphics[scale=.90]{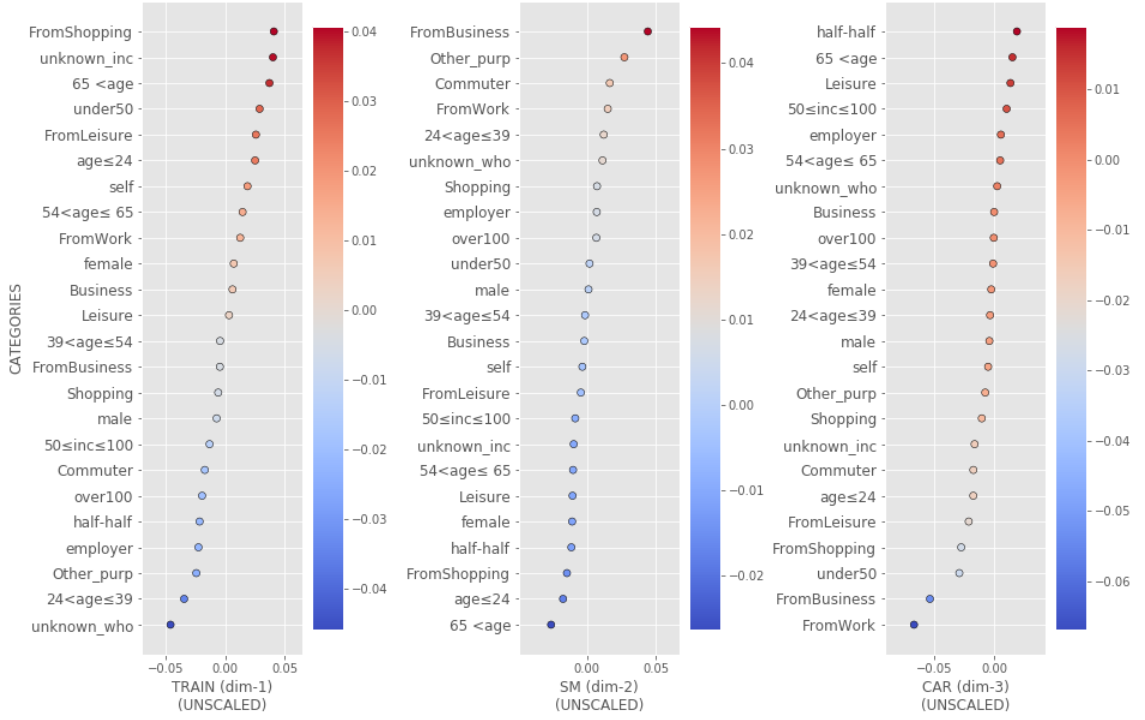}

\caption{\textit{Visualization of the embeddings dimensions per choice alternative for PURPOSE, WHO, AGE, MALE, and INCOME variables in swissmetro dataset.}}
\label{fig:comparison}
\end{figure}

It is also interesting to visualize the ordered embedding values for each mode choice without beta scaling. This would allow us to highlight which categories are not relevant for each mode regardless their associated beta value. It is expected that the categories that have a trivial or no effect on choosing a certain mode will have a close to zero value in its corresponding embedding dimension.
We can observe that the variable MALE has a close to zero value across all the alternative specific embedding dimensions for all the categories it contains (except for female in the SM-specific dimension)\footnote{The embedding vectors for the male and female categories before scaling are $[-0.007,  0.001, -0.004]$ and $[ 0.007, -0.011, -0.002]$ respectively.}, and thus it has an overall negligible effect on choosing any of the modes in the choice set. Interestingly MALE variable has been identified as statistically insignificant by all the proposed models (see Table~\ref{tab:variab}). We note that a similar trend is shown for the variables GA and SEATS \footnote{The embedding vectors (before scaling) for the GA and NO\_GA categories are $[ 0.004,  0.005, -0.004]$ and $[-0.001, 0.002, 0.004]$, while the ones for AIRLINE\_SEATS and NO\_AIRLINE\_SEATS are $[0.022, -0.019, -0.002]$ and $[-0.01, 0.006, -0.003]$ respectively.} that have been identified as statistically insignificant in the best performing model in Table ~\ref{tab:variab}.
These findings suggest that the proposed models are able to isolate from the modeling process not only the categories that have no predictive power by assigning to them embedding values that are close to zero, but also to automatically single out insignificant variables  by assigning to their respective categories very low embedding values. Thus, the embedding representations generated by the proposed models, can be further used diagnostic purposes such as automatic model specification and feature selection. 


Lastly, it is worth-noticing that the embeddings-encoding allows to capture non-linear relationships that may exist
between the target variable and ordinal discrete explanatory variables such as income and age. Indeed, if we observe the magnitudes for the levels corresponding to INCOME (\textit{under50} \textit{50$\leq$inc$\leq$100}, \textit{over100}), across the 3 embedding dimensions, we can observe that their ordering is preserved only in the embedding dimension corresponding to Train. A similar trend is also noticed for the variable AGE that its corresponding levels neither linearly or orderly arranged for all the 3 alternative-specific embedding dimensions.

\newpage
\section{Embeddings: Transferability and Reusability}

In this section, we will focus on one of the advantageous properties of embeddings representations, their reusability, i.e. that the embeddings learned by training the suggested models on one dataset can be reused for other ones, as long as the modeled behavior keeps consistent between the datasets. Training word embeddings on large datasets and saving them with the purpose of reusing them to solve similar tasks in the future is a common practice in the field of NLP, and is considered as a form of Transfer Learning, i.e. using information gained for one task to solve other related ones even if they have different distributions or features \cite{transfer}. Reusing the embeddings representations can be beneficial in terms of computational time and cost and resources,  as well as improving predictive performance in cases of limited or sparse datasets. In this section we will focus on showcasing the aspect of  embeddings \textit{reusability} (reusing the embeddings trained to solve a certain task for \textit{solving the same task} on a different dataset) leaving the aspect of \textit{transferability}  (reusing the embedding trained to solve a certain task for \textit{solving a similar} task on a different dataset) as a future work.


The E-MNL and EL-MNL models presented in this paper aim to offer a NN-based implementation of a MNL that, along with the embeddings generation, allows us to estimate
the utilities coefficients and their post-estimation statistics without the need of using external estimation packages. However, the embeddings representations that are generated from the suggested models can be saved and used for encoding categorical variables -as an alternative to dummy encoding- which can be given as an input to DCM estimation packages of our choice.

\begin{figure}[H]
\includegraphics{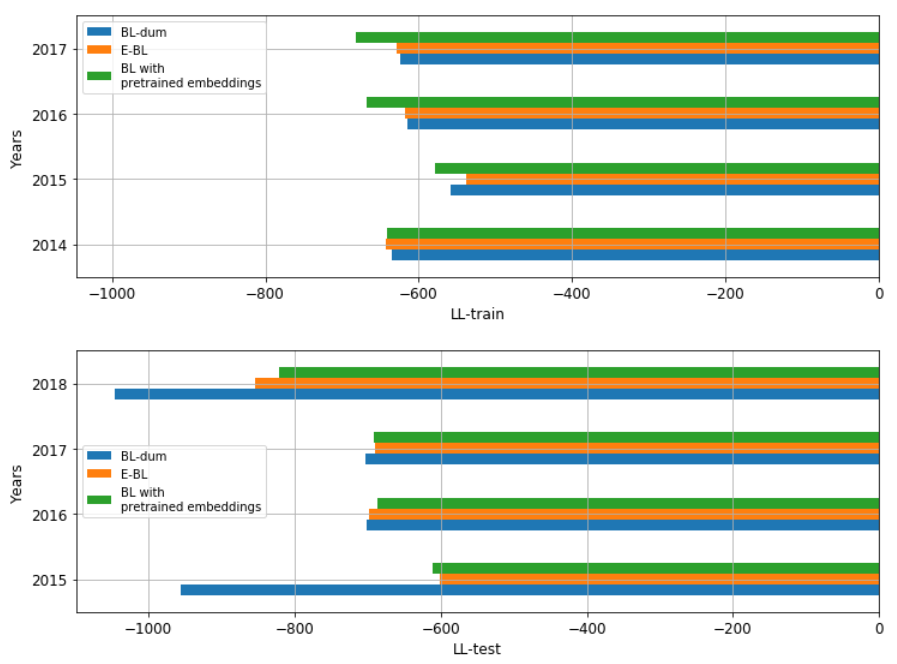} 
\caption{\textit{Comparison of $LL_{train}$ and $LL_{test}$ across the years 2014-2018 between MNL$_{dum}$, E-MNL, and a simple BL with embeddings encoding for the categorical TU variables using E-MNL with pretrained embeddings from year 2014. }}
\label{fig:reuse}
\end{figure}

In order to demonstrate the reusability of the embeddings, we saved a set of embeddings representations that were trained on TU data for the year 2014 (see subsection \ref{tu_experiment}), and used them to encode the categorical variables for the data of the remaining years 2015-2018. Then, we estimated the coefficients for different MNL models for on each of the years 2014-2017,
and tested their predictive performance on the data for the following year in each case. The results are presented in Figure~\ref{fig:reuse} together with the ones for E-MNL and EL-MNL (same values as the ones reported in Table~\ref{tab: TU results} for comparison purposes).

We can observe that, although the MNL model with pretrained embeddings performs worse on the training data compared to the MNL$_{dum}$ model, its predictive performance on the test data is consistently better for all the years considered, showing an improvement between $\sim 3.8\%$ to $27.3\%$.
When comparing the MNL model that uses pretrained embeddings to the E-MNL model (where the embeddings are estimated anew every time on the training data) we can observe that despite the fact that the latter tends to perform better on the training data, they display an overall similar LL on the test sets. Specifically for the years 2015-2017 their absolute difference in LL$_{test}$is ranging between $\sim 0.21$ to $1.76\%$, while for the year 2018 it raises up to $\sim 4\%$ with the E-MNL model performing better. 
These findings indicate that the embeddings trained on $2014$ TU data are reusable for the following years' without substantially compromising the model's predictive performance and therefore their estimation can be viewed as an one-time task.



\section{Discussion and Future Perspectives}

One main direction for future research is to leverage our models' flexible architecture and ability for dimensionality reduction and  expand their architecture to employ heterogeneous data sources for modeling travel behavior, such as text or image-based data that theory-driven DCMs are not yet capable of handling due to their unstructured nature and high dimensionality. These could include for example, responses to open or semi-open questions as in \cite{babu} and \cite{glerum}, mining social media data as in \cite{gu}, or exploiting large crowdsourced datasets, such as StreetScore \cite{streetscore} or Place Pulse \cite{placepulse}, as in \cite{rossetti}, that use Google Street View images to map visual perceptions of public spaces. Additionally, it is worth-noting that although this data is usually available in high volumes, the proposed models would be able to support such inputs efficiently as they use GPU to accelerate the training process. 

Although the reusability of our embeddings 
has been showcased in this study, the aspect of transferability has been left as an open question for future work. Another interesting direction thus, is to explore the potential of Transfer Learning by using the embedding representations trained for solving a given choice task as features for solving similar modeling tasks. For example, using the embeddings that were trained for predicting travel mode choice for making route choice predictions or forecasting departure times. We can however already identify that some challenges and issues may arise along this direction due to the fact that the proposed embedding representations are produced by a single supervised process. As a result, the relations between the categories that are captured by the embeddings are task-dependent, and their transferability to new contexts and tasks may be limited. Thus, although the transferability limits of such representations are worth-investigating, one cannot yet expect to use the proposed (single-output) models for creating pre-trained universal embeddings collections that can be used in a wide variety of predictive tasks, as is the case in the NLP field with word embeddings \footnote{In the field of NLP the creation and use of gold-standard datasets of pretrained word embeddings is a common practice.}. 

As a first step towards this direction, we intend to modify the architecture of the proposed models to support multiple outputs, following an approach that in machine learning literature is referred to as Multi-task learning (MTL), which involves the simultaneous learning of multiple tasks by a shared model \cite{multitask}. MTL leverages shared information among different datasets and similarities across relevant tasks to extract common features and enhance the performance of individual tasks \cite{multitask, multitask2}. This approach would ultimately allow us to obtain more generalized embeddings representations learned across different datasets and target variables that contain rich contextual information and aiming to describe travel-behavior as a whole.

Due to the fact the primary purpose of this paper is to introduce a novel method of encoding discrete variables within the logit framework, and to demonstrate how they can be used for prediction and analysis purposes, the proposed models are limited to cover the standard MNL formulation. Therefore the adaptation of E-MNL and EL-MNL to more sophisticated formulations, such as the Nested Logit, the Mixed Logit  and the Latent Class models and can be considered as future work. It is worth-noting that Sifringer et al. in \cite{sin} implemented a nested generalization of their proposed L-MNL model, the Learning Multinomial Nested Logit (L-NL). In their study they reported that an increase in fit lowers the need of using a nested structure, indicating that the data-driven component allowed for reducing the misspecification bias that was responsible for the initially observed correlation
between alternatives. Thus, it would be interesting to apply the suggested E-MNL and EL-MNL models to more advanced DCM formulations and explore whether the latent embedding representations are able to sufficiently account for noise and unobserved heterogeneity and to which extent they can reduce the need of deploying more complex theory-driven models. 

Lastly, in addition to the code and practical examples that accompany this paper, we intend to develop a Python package with the implementation of the proposed (and future) models and integrate it as an independent application, hoping that it will promote and contribute to the ongoing research towards the understanding and modeling of traveling behavior.

\section{Conclusions}

This study builds on previous work that combines theory-driven and data-driven choice models and introduces an ANN-based DCM that uses an embedding layer to project categorical variables into a lower dimensional latent space. Specifically, we proposed a basic (E-MNL) and an extended (EL-MNL) model formulation within a flexible framework that estimates jointly the linear and non-linear models' parameters, and, in contrast to previous studies, provides utility coefficients and post-estimation statistics for all the input variables. 
Both models were tested on two real world datasets and evaluated against benchmark and baseline models in terms of predictive performance, interpretability, model complexity and transparency.
In contrast to previous studies that used embedding representations within the logit framework, the novelty of our work lies in the learning of latent yet interpretable representations by formally associating each embedding dimension to a choice alternative. The results indicate that the suggested embedding encoding allows us to
produce compact representations that provide behaviorally meaningful outputs which can be used for diagnostic and analysis purposes, such as automatic model specification, feature selection, and pattern recognition. These rich representations lead to more parsimonious, and transparent models that are beneficial in terms of predictive performance and more robust to overfitting compared to previously suggested ANN-based DCMs and simple baseline DCMs that use dummy-encoding. Additionally, it was shown that the extended EL-MNL architecture can improve predictive performance compared to the simpler E-MNL. However we recommend that its use should be assessed based on the complexity of the dataset and the problem at hand, as the E-MNL alone may suffice to obtain the same performance levels using less number of parameters. Our code is publicly available so that future researchers can reproduce our results, expand our work, and perform their own experiments on other domains and applications.

\end{document}